\documentclass[10pt,twocolumn]{IEEEtran}
\usepackage{graphicx}
\usepackage{epsfig}
\usepackage{subfigure}
\usepackage{amssymb}
\usepackage{amsbsy}
\usepackage{amsmath}
\usepackage{cite}

\newcounter{theorem}
\newtheorem{theorem}{Theorem}
\newtheorem{lemma}[theorem]{Lemma}
\newtheorem{claim}[theorem]{Claim}
\newcounter{definition}
\newtheorem{definition}{Definition}

\newcommand{\beq}{\begin{equation}}
\newcommand{\eeq}{\end{equation}}
\newcommand{\bea}{\begin{array}}
\newcommand{\ena}{\end{array}}
\newcommand{\bds}{\begin {itemize}}
\newcommand{\eds}{\end {itemize}}
\newcommand{\bdf}{\begin{definition}}
\newcommand{\blm}{\begin{lemma}}
\newcommand{\edf}{\end{definition}}
\newcommand{\elm}{\end{lemma}}
\newcommand{\bthm}{\begin{theorem}}
\newcommand{\ethm}{\end{theorem}}
\newcommand{\bprp}{\begin{prop}}
\newcommand{\eprp}{\end{prop}}
\newcommand{\bcl}{\begin{claim}}
\newcommand{\ecl}{\end{claim}}
\newcommand{\bcr}{\begin{coro}}
\newcommand{\ecr}{\end{coro}}
\newcommand{\bquest}{\begin{question}}
\newcommand{\equest}{\end{question}}

\newcommand{\larrow}{{\larrow}}

\newcommand{\nin}{{\not \in}}

\def\urltilda{\kern -.15em\lower .7ex\hbox{\~{}}\kern .04em}

\begin{document}\title{Learning in Restless Multi-Armed Bandits via Adaptive Arm Sequencing Rules}
\author{Tomer Gafni and Kobi Cohen
\thanks{Tomer Gafni and Kobi Cohen are with the School of Electrical and Computer Engineering, Ben-Gurion University of the Negev, Beer-Sheva 84105, Israel. Email: gafnito, yakovsecg@bgu.ac.il}
\thanks{This work has been submitted to the IEEE for possible publication. Copyright may be transferred without notice, after which this version may no longer be accessible.}
\thanks{A short version of this paper was presented at IEEE International Symposium on Information Theory (ISIT) 2018 [1]}
}
\date{}
\maketitle

\begin{abstract}
\label{sec:abstract}
We consider a class of restless multi-armed bandit (RMAB) problems with unknown arm dynamics. At each time, a player chooses an arm out of $N$ arms to play, referred to as an active arm, and receives a random reward from a finite set of reward states. The reward state of the active arm transits according to an unknown Markovian dynamics. The reward state of passive arms (which are not chosen to play at time $t$) evolves according to an arbitrary unknown random process. The objective is an arm-selection policy that minimizes the regret, defined as the reward loss with respect to a player that always plays the most rewarding arm. This class of RMAB problems has been studied recently in the context of communication networks and financial investment applications. We develop a strategy that selects arms to be played in a consecutive manner, dubbed Adaptive Sequencing Rules (ASR) algorithm. The sequencing rules for selecting arms under the ASR algorithm are adaptively updated and controlled by the current sample reward means. By designing judiciously the adaptive sequencing rules, we show that the ASR algorithm achieves a logarithmic regret order with time, and a finite-sample bound on the regret is established. Although existing methods have shown a logarithmic regret order with time in this RMAB setting, the theoretical analysis shows a significant improvement in the regret scaling with respect to the system parameters under ASR. Extensive simulation results support the theoretical study and demonstrate strong performance of the algorithm as compared to existing methods.
\end{abstract}
%
\section{Introduction}
\label{sec:introduction}

Restless Multi-Armed Bandit (RMAB) problems are generalizations of the classic Multi-Armed Bandit (MAB) problem \cite{Gittins_1979_Bandit, Lai_1985_Asymptotically, Anantharam_1987_Asymptotically}. Differing from the classic MAB, where the states of passive arms remain frozen, in the RMAB setting, the state of each arm (active or passive) can change. The RMAB problem under the Bayesian formulation with known Markovian dynamics has been shown to be P-SPACE hard in general \cite{Papadimitriou_1999_Complexity}.

In this paper, we consider the following RMAB problem with unknown arm dynamics. At each time, a player chooses an arm out of $N$ arms to play, referred to as an active arm. Once playing an arm, a random reward is received from a finite set of rewards. The reward state of the active arm transits according to an unknown Markovian dynamics. The reward state of passive arms (which are not chosen to play at time $t$) might change as well and evolve according to an arbitrary unknown random process.

This class of RMAB problems has been studied recently in the context of communication networks, and financial investment applications \cite{Tekin_2012_Online, Liu_2013_Learning}. For example, in the hierarchical opportunistic spectrum access model in cognitive radio networks, a secondary user (unlicensed) is allowed to transmit data over a channel among a set of available channels (i.e., arms) when primary (licensed) users do not transmit. The temporal spectrum usage of the primary user is modeled by a Markovian dynamics, which leads to a Markovian reward model. Thus, the secondary user aims at designing a good channel selection policy without knowing the dynamics of the primary users, with the goal of maximizing its long-term rate (i.e., accumulated reward). Other related models have studied channel selection under unknown fading dynamics and financial investments (see \cite{Tekin_2012_Online, Liu_2013_Learning} and references therein).

\subsection{Performance Measures of Learning in RMAB}
\label{ssec:intro_performance}

Although optimal solutions have been obtained for some special cases of RMAB models (see references in Section \ref{ssec:related}), solving RMAB problems directly is intractable in general \cite{Papadimitriou_1999_Complexity}. Thus, a widely used performance measure of an algorithm is the \emph{regret}, defined as the reward loss with respect to a player with a side information on the model. An algorithm that achieves a sublinear scaling rate of the regret with time approaches the performance of the player with the side information as time increases. The essence of the problem is thus to design an algorithm that learns the side information effectively so as to achieve the best sublinear scaling of the regret with time.

In this paper we use the definition of regret that was introduced in \cite{Auer_2002_Nonstochastic} and used later in \cite{Tekin_2012_Online, Liu_2013_Learning} for a similar RMAB model as considered here. Specifically, the regret is defined as the reward loss of an algorithm with respect to a player that knows the expected reward of all arms and always plays the arm with the highest expected reward. It should be noted that computing the optimal policy for RMABs is P-SPACE hard even when the Markovian model is known \cite{Papadimitriou_1999_Complexity}. Nevertheless, always playing the arm with the highest expected reward is known to be optimal in the classic MAB under i.i.d. or rested Markovian rewards (up to an additional constant term \cite{Anantharam_1987_Asymptotically}). Thus, it is commonly used in RMAB with unknown dynamics settings for measuring the algorithm performance in a tractable manner.

\subsection{Existing Random and Deterministic Approaches}
\label{sec:existing}

We are facing an online learning problem with the well known exploration versus exploitation dilemma. On the one hand, a player should explore all arms in order to infer their states. On the other hand, it should exploit the information gathered so far to play the best arm. Due to the restless nature of both active and passive arms and potential reward loss due to transient effect as compared to steady state when switching arms, learning the Markovian reward statistics requires that arms will be played in a consecutive manner for a period of time (i.e., epoch).
In \cite{Tekin_2012_Online, Liu_2013_Learning}, regenerative cycle algorithm (RCA), and deterministic sequencing of exploration and exploitation (DSEE) algorithm, respectively, have been proposed based on these insights. The RCA algorithm chooses the active arms based on the upper confidence bound (UCB) index \cite{auer2002finite} when entering each epoch, and a logarithmic regret with time was shown. However, since RCA performs random regenerative cycles until catching predefined states at each epoch (i.e., hitting times) the scaling with the mean hitting time $M$ (which scales at least polynomially with the state space) is of order $O(M\log t)$. The DSEE algorithm overcomes this issue by using deterministic sequencing of exploration and exploitation epochs. A logarithmic regret with time was shown under DSEE. However, applying the deterministic sequencing method by DSEE results in oversampling bad arms to achieve the desired logarithmic regret, which scales as $O\left((\frac{1}{\sqrt{\Delta}}+\frac{N-2}{\Delta})\log t\right)$, where $N$ is the number of arms and $0<\Delta<(\mu_{\sigma(1)}-\mu_{\sigma(2)})^2$ is a known lower bound on the square difference between the highest reward mean $\mu_{\sigma(1)}$ and the second highest reward mean $\mu_{\sigma(2)}$. Increasing the mean hitting times (e.g., by increasing the state space, or decreasing the probability of switching between states) decreases performance under RCA. Increasing $N$ when $(\mu_{\sigma(1)}-\mu_{\sigma(2)})$ is small as compared to the differences between $\mu_{\sigma(1)}$ and the reward means of other arms decreases performance under DSEE.

\subsection{Main Results}

Our main results are summarized below: \vspace{0.2cm}
\begin{enumerate}
\item \emph{Algorithm development:} We propose a novel Adaptive Sequencing Rules (ASR) algorithm for solving the RMAB problem. The basic idea of ASR is to estimate online the desired (unknown) exploration rate of each arm required for efficient learning. Thus, by sampling each arm according to the desired exploration rate, ASR avoids oversampling bad arms as in DSEE, and at the same time it avoids using too frequent regenerative cycles as in RCA. Interestingly, the size of the exploitation epochs is deterministic and the size of the exploration epochs is random under ASR. The sequencing rules that decide when to enter each epoch are adaptive in the sense that they are updated dynamically and controlled by the current sample means in a closed-loop manner. \vspace{0.2cm}
\item \emph{Theoretical performance analysis:} We establish a finite-sample upper bound on the regret under the proposed ASR algorithm. Our analysis is valid for both model settings in \cite{Tekin_2012_Online}, and \cite{Liu_2013_Learning}. Thus, performance comparison between the algorithms can be conducted analytically. Specifically, similar to RCA \cite{Tekin_2012_Online} and DSEE \cite{Liu_2013_Learning}, we show that the proposed ASR algorithm achieves a logarithmic regret order with time as well. The scaling with the mean hitting time under ASR, however, is significantly better than the scaling under RCA ($O(M\log\log t)$ under ASR as compared to $O(M\log t)$ under RCA). The scaling with the number of arms and $\Delta$ under ASR is significantly better than the scaling under DSEE ($O\left((\frac{1}{\sqrt{\Delta}}+N-2)\log t\right)$ under ASR as compared to $O\left((\frac{1}{\sqrt{\Delta}}+\frac{N-2}{\Delta})\log t\right)$ under DSEE).\vspace{0.2cm}
\item \emph{Simulation results:} We performed extensive simulation experiments that support our theoretical results under various parameter settings. Significant performance gain of ASR over RCA and DSEE has been observed.
\end{enumerate}

\subsection{Related Work}
\label{ssec:related}

RMAB problems have been studied under both the non-Bayesian \cite{Dai_2011_Non, Tekin_2012_Online, Tekin_2012_Approximately, Liu_2013_Learning, oksanen2015order} and Bayesian \cite{Whittle_1988_Restless, Weber_1990_On, Ehsan_2004_On, Zhao_2008_Myopic, Ahmad_2009_Optimality, Ahmad_2009_Multi, Liu_2010_Indexability, Wang_2012_Optimality, wang2014optimality} settings.
Under the non-Bayesian setting, special cases of Markovian dynamics have been studied in \cite{Dai_2011_Non, Tekin_2012_Online, oksanen2015order}. Under the Bayesian setting with known dynamics, the objective is exact optimality in terms of the total expected reward over time. The structure of the optimal policy for a general RMAB remains open. There are a number of studies that focused on special classes of RMABs. In particular, the optimality of the myopic policy was shown under positively correlated two-state Markovian arms \cite{Zhao_2007_Structure,Zhao_2008_Myopic, Ahmad_2009_Optimality, Ahmad_2009_Multi} under the model where a player receives a unit reward for each arm that was observed in a good state. In \cite{Liu_2010_Indexability, Liu_2011_Indexability}, the indexability of a special classes of RMAB has been established. In \cite{Wang_2012_Optimality}, a family of regular reward functions, satisfying axioms of symmetry, monotonicity and decomposability, was introduced and optimality conditions of a myopic policy have been established under that family of reward functions. The authors have generalized the results under imperfect sensing in \cite{wang2014optimality}. In our previous work, optimality conditions of a myopic policy have been derived under arm activation constraints \cite{cohen2014restless}.

Other related work considered the RMAB formulation under the compressive spectrum sensing problem in cognitive radio networks \cite{bagheri2015restless}, where the focus was on deriving the myopic policy. In \cite{Koutsopoulos_2010_Optimal}, the problem of transmission rate control was cast as POMDP, where an optimal threshold policy was established under slowly time-varying link with two states. In \cite{yu2018deadline} the stochastic deadline scheduling problem was formulated as a restless multi armed bandit problem. Other recent extensions of the MAB problem can be found in \cite{lesage2017multi, reverdy2017satisficing, meshram2018whittle}. In \cite{lesage2017multi}, the authors considered the case in which the number of active arms evolves as a stationary process. In \cite{reverdy2017satisficing}, the MAB problem was defined with satisfying objectives so that the player aims at obtaining a reward above a certain threshold. In \cite{meshram2018whittle}, a two states RMAB problem was defined in a hidden nature. Specifically, when an arm is sampled, the state of the arm is not fully observable. Instead, a random binary signal is received that depends on the state of the arm. However, the RMAB model considered in this paper is fundamentally different from these studies.
Other related approaches include game theoretic, and reinforcement learning algorithms (see \cite{cohen2013game,cohen2015distributed,
cohen2017distributed,naparstek2017deep,naparstek2019deep} and references therein).
\section{System Model and Problem Formulation}
\label{sec:problem}

We consider $N$ arms indexed by $i=1,2,\cdots,N$. The $i^{th}$ arm is modeled as a discrete-time, irreducible, aperiodic and reversible Markov chain with finite state space $S^i$.
At each time, the player chooses one arm to play. Each arm, when played, offers a certain positive reward that defines the current state of the arm. Let $s_i(t)$ denote the state of arm $i$ at time $t$, and
\begin{center}
$\displaystyle s_{max}\triangleq \max_{s \in S^i, 1 \leq i \leq N } s, \quad r_{max}\triangleq\max_{1\leq i \leq N}\sum_{s\in{S^i}}s $
\end{center}
\begin{center}
$ \displaystyle S_{max} \triangleq \max_{1 \leq i \leq N}|S^i| $
\end{center}

Let $P^i$ denote the transition probability matrix and {$\vec{\pi_i}$} $=\{\pi_i^s\}_{s\in{S^i}}$ be the stationary distribution of arm $i$. We define
\begin{center}
$\displaystyle\pi_{min}\triangleq\min_{1\leq i \leq N,s \in S^i}\ \pi_i^s, \quad
\hat{\pi}_i^s \triangleq \max\{\pi_i^s,1-\pi_i^s\}   $.
\end{center}
\begin{center}
$ \displaystyle \hat{\pi}_{max} \triangleq \max _{s \in S^i, 1 \leq i \leq N } \{\pi_i^s,1-\pi_i^s \}$
\end{center}

Let $\lambda_i$ be the second largest eigenvalue of $P^i$, and let
\begin{center}
$\displaystyle\lambda_{max}\triangleq\max_{1\leq i \leq N}\ \lambda_i$
\end{center}
be the maximal one among all arms. Also, let
\begin{center}
$\displaystyle\overline{\lambda}_{min}\triangleq 1-\lambda_{max}$,
\end{center}
and let
\begin{center}
$\displaystyle\overline{\lambda_i}\triangleq 1-\lambda_i$
\end{center}
be the eigenvalue gap.

Let $M^i_{x,y}$  be the mean hitting time of state $y$ starting at initial state $x$ for arm $i$, and let
\begin{center}
$\displaystyle M^i_{max}\triangleq\max_{x,y \in S^i, x\neq y}M^i_{x,y}$.
\end{center}
We also define:
\beq
\label{eq:L}
\bea{l}
\displaystyle A_{max}\triangleq\max_{i}\;(\min_{ s \in S^i}\ \pi_i(s))^{-1} \sum\limits_{s\in{S^i}}s,\vspace{0.2cm}\\
\displaystyle L\triangleq\frac{30r_{max}^2}{(3-2\sqrt{2})\overline{\lambda}_{min}}.\vspace{0.0cm}\\
\ena
\eeq

We assume that the arms are restless. Specifically, the reward state of the active arm (say $i$) transits according to the unknown Markovian rule $P^i$, while the reward state of passive arms (which are not chosen to play at time $t$) evolves according to an arbitrary unknown random process.
The stationary reward mean $\mu_i$ is given by
\begin{center}
$\displaystyle\mu_i=\sum\limits_{s\in{S^i}}s\pi_i(s)$.
\end{center}
Let $\sigma$ be a permutation of $\{1,...,N\}$ such that
\begin{center}
$\mu^*\triangleq\mu_{\sigma(1)}\geq\mu_{\sigma(2)}\geq \cdots\geq \mu_{\sigma(N)}$.
\end{center}
Let $t^i(n)$ denote the time index of the $n^{th}$ play on arm $i$, and $T^i(t)$ denote the total number of plays on arm $i$ by time $t$.
Thus, the total reward by time $t$ is given by:
\beq
\displaystyle R(t)= \sum\limits_{i=1}^{N}\sum\limits_{n=1}^{T^i(t)}s_i(t^i(n)).
\eeq

Let $\phi(t)\in\left\{1, 2, ..., N\right\}$ be a selection rule indicating which arm is chosen to be played at time~$t$, which is a mapping from the observed history of the process (i.e., all past actions and observations up to time $t-1$) to $\left\{1, 2, ..., N\right\}$ (can also be a randomized selection that maps to a probability mass function over selected arms). A policy $\phi$ is the time series vector of selection rules: $\phi=(\phi(t), t=1, 2, ...)$. As explained in Section \ref{ssec:intro_performance}, we define the regret $r^\phi(t)$ for policy $\phi$ as the difference between the expected total reward that can be obtained by playing the arm with the highest mean, and the expected total reward obtained from using policy $\phi$ up to time $t$:
\beq
r_\phi(t)= t\mu_{\sigma(1)}-\mathbb{E}_\phi[R(t)].
\eeq
The objective is to find a policy that minimizes the growth rate of the regret with time. In the next section we propose the Adaptive Sequencing Rules (ASR) Algorithm to achieve this goal.

\section{The Adaptive Sequencing Rules (ASR) Algorithm}
\label{sec:ASR}

The basic idea of the ASR algorithm is to sample each arm according to its exploration rate required for efficient learning. We show in the analysis that we must explore a bad arm $\sigma(i)$, $i=2, 3, ..., N$, at least $\overline{D}_i\log t$ times for being able to distinguishing it from $\mu^*$ with a sufficiently high accuracy, where
\beq
\label{eq:Dbar}
\overline{D}_i\triangleq{\frac{4L}{(\mu^*-\mu_{\sigma(i)})^2}}.
\eeq
The smaller the mean difference, the more samples we must take for exploring bad arms.
\emph{Since the reward means $\left\{\mu_{\sigma(i)}\right\}_{i=1}^N$, are unknown, however, we can estimate $\overline{D}_i$ by replacing $\mu_{\sigma(i)}$ by its sample reward mean. Using the estimate of $\overline{D}_i$ (which is updated dynamically during time and controlled by the sample means), we can design an adaptive sequencing rule for sampling arm $i$ that will converge to its exploration rate, required for efficient learning, as time increases.} Whether we succeed to obtain a logarithmic regret order depends on how fast the estimate of $\overline{D}_i$ converges to a value which is no smaller than $\overline{D}_i$ (so that we take at least $\overline{D}_i$ samples from bad arms in most of the times). To guarantee the desired convergence speed, we judiciously overestimate $\overline{D}_i$ as detailed in Section \ref{ssec:selection}.  \vspace{0.2cm}

\begin{figure*}[t]
\centering \epsfig{file=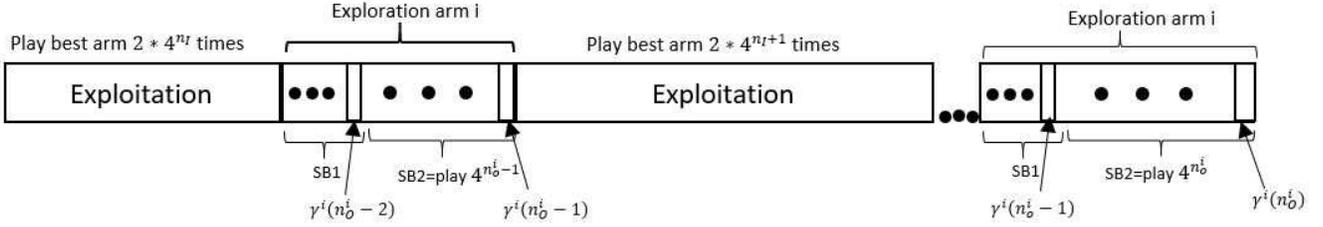,
width=1.0\textwidth}
\caption{An illustration of the exploration and exploitation epochs under ASR. As explained in Section \ref{ssec:exploitation}, during an exploitation epoch the player plays the same arm that had the highest sample mean in the beginning of the epoch. As explained in Section \ref{ssec:exploration}, an exploration epoch is divided into a random-size sub-block SB1 and a deterministic (geometrically growing) size sub-block SB2. SB1 of an arm (say $i$ as in the figure) is a random hitting time until catching the last state $\gamma^i$ that arm $i$ observed in the previous exploration epoch. Selecting which type of epoch to play is determined by the selection rule described in Section \ref{ssec:selection}.}
\label{fig:epochs}
\end{figure*}

\subsection{Playing arms consecutively during exploration and exploitation epochs:}
\label{ssec:interleaving}
As discussed in Section \ref{sec:existing}, learning the Markovian reward statistics requires playing arms in a consecutive manner for a period of time. For instance, the RCA algorithm selects arms based on the UCB index and plays the arm for a random period of time which depends on hitting time events. On the other hand, the DSEE algorithm samples arms for a deterministic periods of time that grow geometrically with time. Interestingly, we show that by judiciously combining these two sampling methods, while setting the exploration rate for each arm according to its adaptive sequencing rule (described in Section \ref{ssec:selection}), we can achieve tremendous improvement in both theoretical and simulation performance as shown in Section \ref{sec:regret}.

Specifically, we divide the time horizon into exploration and exploitation epochs, as illustrated in Fig. \ref{fig:epochs}. An exploration epoch is dedicated to play a certain arm determined by its adaptive sequencing rule (described in Sections \ref{ssec:selection}, \ref{ssec:pseudocode}). We define $n_O^i(t)$ as the number of exploration epochs in which arm $i$ was played up to time $t$. An exploitation epoch is dedicated to play the arm with the highest sample mean, whenever exploration is not executed. We define $n_I(t)$ as the number of exploitation epochs up to time $t$. In Fig. \ref{fig:epochs}, we illustrate the exploration epochs for arm $i$ only, for the ease of illustration. In general, an interleaving of exploration epochs for all arms with exploitation epochs (for the arm with the highest sample mean) is performed. \vspace{0.2cm}

\subsection{The structure of exploration epochs:}
\label{ssec:exploration}

The exploration epochs for each arm are divided into two sub-blocks: a random-size sub-block SB1, and a deterministic-size sub-block SB2. Consider time $t$ (and we remove the time index $t$ for convenience). Let $\gamma^i(n_O^i-1)$ be the last reward state observed at the $(n_O^i-1)^{th}$ exploration epoch. As illustrated in Fig. \ref{fig:epochs}, once the player starts the $(n_O^i)^{th}$ exploration epoch, it first plays a random period of time until observing $\gamma^i(n_O^i-1)$ (i.e., a random hitting time). This random period of time is referred to as SB1. Then, the player plays a deterministic period of time with length $4^{n_O^i}$. This deterministic period of time is referred to as SB2. The player stores the last reward state $\gamma^i(n_O^i)$ observed at the current $(n_O^i)^{th}$ exploration epoch, and so on. We define the set of time indices during SB2 sub-blocks by $\mathcal{V}_i$
\vspace{0.2cm}

\subsection{The structure of exploitation epochs:}
\label{ssec:exploitation}
Let $\overline{s}_i$ be the sample reward mean of arm $i$ when entering the $(n_I)^{th}$ exploitation epoch. Then, the player plays the arm with the highest sample mean $\max_i \overline{s}_i$ for a deterministic period of time with length $2\cdot4^{n_I-1}$ (there are no arm switchings inside epochs). We define the set of time indices in the exploitation epochs by $\mathcal{W}_i$. Computing the sample mean $\overline{s}_i$ for each arm is based on observations taken from $\mathcal{V}_i$ and $\mathcal{W}_i$. Observations from SB1 sub-blocks are removed to ensure the consistency of the estimators.\vspace{0.2cm}

\subsection{The Selection rule (choosing between epoch types):}
\label{ssec:selection}

At the beginning of each epoch, the player needs to decide whether to enter an exploration epoch for one of the $N$ arms, or whether to enter an exploitation epoch for the arm with the highest sample mean. Let $\widetilde{s}_i(t)$ be the sample reward mean of arm $i$, computed based on observations taken from $\mathcal{V}_i$ only\footnote{Since the algorithm creates continuity in the sample series taken from $\mathcal{V}_i$, we show in the analysis that $\widetilde{s}_i$ used in the computation of $\widehat{D}_i(t)$ enables a sufficiently fast convergence to the desired exploration rate.} at time $t$. Let
\beq
\label{eq:Dhat}
\widehat{D}_i(t)\triangleq\frac{4L}{\max\left\{\Delta, (\max_{j}\widetilde{s}_j(t)-\widetilde{s}_i(t))^2-\epsilon\right\}},
\eeq
where $0<\Delta<(\mu_{\sigma(1)}-\mu_{\sigma(2)})^2$ is a known lower bound on the square difference $(\mu_{\sigma(1)}-\mu_{\sigma(2)})^2$, and $\epsilon>0$ is a fixed tuning parameter (in practice, these parameters are not needed and only used for theoretical analysis, as discussed in Section \ref{ssec:pseudocode}).
We also define:
\beq
\label{eq:I}
\displaystyle I\triangleq\frac{\epsilon^2 \cdot \overline{\lambda}_{min}}{192(r_{max}+2)^2 \cdot S_{max}^2 \cdot r_{max}^2 \cdot \hat{\pi}_{max}^2}.
\eeq

We next design the selection rule based on the following insights. First, the algorithm must take at least $\overline{D}_i\log t$ samples from each bad arm ($\overline{D}_i$ is given in (\ref{eq:Dbar})) for computing a sufficiently accurate sample means $\overline{s}_i$. Therefore, the algorithm replaces the unknown value $\overline{D}_i$ by $\widehat{D}_i(t)$, which overestimates $\overline{D}_i$ to obtain the desired property.
Second, since $\widehat{D}_i(t)$ is a random variable, we need to make sure that the desired property holds with a sufficiently high probability. The term $I$ in (\ref{eq:I}) can be viewed as the minimal rate function of the estimators among all arms and is used to guarantee the desired property (note that decreasing $\epsilon$ decreases $I$, which increases the required sample size).
Consider a beginning of each epoch at time $t$, and let $\mathcal{V}_i(t)$ be the set of all time indices during SB2 sub-blocks up to time $t$. Then, if there exists an arm (say $i$) such that the following condition holds:
\beq
\label{eq:condition}
|\mathcal{V}_i(t)|\leq\max\left\{\widehat{D}_i(t), \frac{2}{I}\right\}\cdot\log t,
\eeq
then the player enters an exploration epoch for arm $i$ (ties between arms are broken arbitrarily). Otherwise, it enters an exploitation epoch. As a result, the selection rule for each arm that governs the arm sequencing policy is adaptive in the sense that it is updated dynamically with time and controlled by the random sample mean in a closed loop manner.  \vspace{0.2cm}

\subsection{High-level pseudocode and implementation of ASR:}
\label{ssec:pseudocode}

In summary, the player performs the following algorithm:\vspace{0.2cm}\\
\noindent 1) (Initialization:) For all $N$ arms, execute an exploration epoch where a single observation is taken from each arm.
\vspace{0.2cm}\\
\noindent 2) If condition (\ref{eq:condition}) holds for some arm (say $i$), then execute an exploration epoch for arm $i$ (as described in Section \ref{ssec:exploration}) and go to Step 2 again. Otherwise, go to Step 3.
\vspace{0.2cm}\\
\noindent 3) Execute an exploitation epoch (as described in Section \ref{ssec:exploitation}) and go to Step 2. \vspace{0.2cm}

We next discuss technical implementation details when executing the ASR algorithm. (i) From a theoretical perspective, ASR and DSEE require the same knowledge on the system parameters to guarantee the theoretical performance. RCA requires the same parameters, except that $\Delta$ is not needed. (ii) It is well known that there is often a gap between the sufficient conditions required by theoretical analysis (often due to union-bounding events in analysis) and practical conditions used for efficient online learning. For example, in \cite{Liu_2013_Learning} the authors simulated DSEE with exploration rate $10\cdot\log t$ while the theoretical sufficient conditions were $\approx 1,000\cdot\log t$. A similar gap was observed in RCA. Indeed, this is the case in ASR as well. While the sufficient conditions provided by the theoretical analysis in Section \ref{sec:regret} require to overestimate $\overline{D}_i$ as in (\ref{eq:Dhat}), simulation results provide much better performance when estimating $\overline{D}_i$ directly by setting $\widehat{D}_i(t)\leftarrow\frac{4L}{(\max_{j}\widetilde{s}_j(t)-\widetilde{s}_i(t))^2}$. Thus, in practice $\Delta$ is not needed and the parameters can be estimated on the fly.

\section{Regret Analysis}
\label{sec:regret}

\subsection{Theoretical Regret Analysis under ASR}
In the following theorem we establish a finite-sample bound on the regret with time. The proof can be found in the Appendix.

\begin{theorem}
\label{th:regret}
Assume that the proposed ASR algorithm is implemented and the assumptions on the system model described in Section \ref{sec:problem} hold. Then, the regret at time $t$ is upper bounded by:
\beq
\bea{l}
\label{eq:regret}
\displaystyle r(t) \leq C_1 \cdot \log_4(t)+C_2\cdot \log(t)\vspace{0.2cm}\\
\hspace{0.0cm}\displaystyle+ \bigg( N \cdot A_{max} + \sum\limits_{i=2}^{N}\left(\mu_{\sigma(1)}-\mu_{\sigma(i)}\right)M_{max}^i \bigg) \cdot \log_4(\log(t))\\
+O(1),
\ena
\eeq
where
\begin{center}
$\bea{l}
\displaystyle
C_1=A_{max}+3\sum_{i=2}^{N}\frac{\mu_{\sigma(1)}-\mu_{\sigma(i)}}{\pi_{min}}\times \vspace{0.2cm}\\
\hspace{2cm}
\displaystyle
\sum\limits_{k=1,i} \left(\frac{1}{\log(2)}+\frac{\sqrt{2}\overline{\lambda}_k \sqrt{L}}{10 \sum\limits_{s\in{S_k}}s}|S_k|\right),
\ena$
\end{center}
\beq
\bea{l}
\displaystyle
C_2=4\sum_{i=2}^{N}\left[\textbf{1}_{\mathcal{K}}(i)\max\left\{(\mu_{\sigma(1)}-\mu_{\sigma(i)})\frac{2}{I}\;,\right.\right.\vspace{0.2cm}\\\hspace{1.5cm}\left.
\displaystyle
\frac{4L}{(\mu_{\sigma(1)}-\mu_{\sigma(i)})+\sqrt{2\epsilon}}+\frac{4L \cdot \sqrt{2\epsilon}}{(\mu_{\sigma(1)}-\mu_{\sigma(i)})^2-2\epsilon} \right\}
\vspace{0.2cm} \\\hspace{2cm}\left.
\displaystyle
+\textbf{1}_{\mathcal{K}^C}(i)\left(\mu_{\sigma(1)}-\mu_{\sigma(i)}\right)\max\left\{\frac{2}{I},\frac{4L}{\Delta}\right\}\right], \vspace{0.2cm} \\
\ena
\eeq
where $\mathcal{K}$ is defined as the set of all indices $i\in{\{2,...,N\}}$ that satisfy:\vspace{0.2cm}
\begin{center}
$\displaystyle (\mu_{\sigma(1)}-\mu_{\sigma(i)})^2-2\epsilon>(\mu_{\sigma(1)}-\mu_{\sigma(2)})^2$,\vspace{0.2cm}
\end{center}
and $\textbf{1}_{\mathcal{K}}(i)$ is the indicator function on the set $\mathcal{K}$, i.e., $\textbf{1}_{\mathcal{K}}(i)=1$ if $i\in\mathcal{K}$ and $\textbf{1}_{\mathcal{K}}(i)=0$ otherwise. $\mathcal{K}^C$ is the complementary set of $\mathcal{K}$.
\end{theorem}

\subsection{Theoretical Comparison with RCA and DSEE:}
\label{ssec:theoretical}

Theorem \ref{th:regret} shows that similar to RCA \cite{Tekin_2012_Online} and DSEE \cite{Liu_2013_Learning}, the regret under ASR has a logarithmic order with time. The scaling with the mean hitting times under ASR, however, is significantly better than the scaling under RCA. Since RCA performs random regenerative cycles until catching predefined states in each epoch, the scaling with the mean hitting times (which scales at least polynomially with the state space) is $O(\sum_i M_{max}^i\log t)$. On the other hand, ASR scales only with $O(\sum_i M_{max}^i\log\log t)$.
The scaling with $N$ and $\Delta$ under ASR is significantly better than the scaling under DSEE. Specifically, the scaling under DSEE is of order $O\left((\frac{1}{\sqrt{\Delta}}+\frac{N-2}{\Delta})\log t\right)$ since all bad arms are explored according to the worst exploration rate. The scaling under ASR, however, is of order\footnote{Note that when $\mu_{\sigma{(1)}}-\mu_{\sigma{(2)}}$ decreases, while other reward means are fixed (i.e., $\Delta$ decreases), we can choose $\epsilon>0$ so that $\mathcal{K}^C$ contains a single bad arm (with the second highest mean). Otherwise, the scaling is of order $O\left((\frac{1}{\sqrt{\Delta}}+\frac{|\mathcal{K}^C|-1}{\Delta}+|\mathcal{K}|)\log t\right)$.}
$O\left((\frac{1}{\sqrt{\Delta}}+N-2)\log t\right)$ since every bad arm is sampled according to its unique exploration rate which is estimated by the adaptive sequencing rules.
\vspace{0.2cm}

\subsection{Numerical Comparison with RCA and DSEE:}

In this section, we analyze the performance of ASR numerically as compared to DSEE and RCA under typical settings of dynamic spectrum access in cognitive radio networks. We simulated a variety of scenarios under the commonly used Gilbert-Elliot channel model. Specifically, each arm is represented by a channel that has two states, \textit{good} and \textit{bad} (or 1,0, respectively). In Section \ref{ssec:state_space} we simulated arms with $20$ states. \emph{In all simulations, we implemented ASR without tuning $\epsilon$ (set to zero), and $\widehat{D}_i(t)$ was estimated on the fly (without the knowledge of $\Delta$), as discussed in Section \ref{ssec:pseudocode}.} \vspace{0.2cm}

\subsubsection{RMAB with $5$ arms}

\begin{figure}[htbp]
\centering \epsfig{file=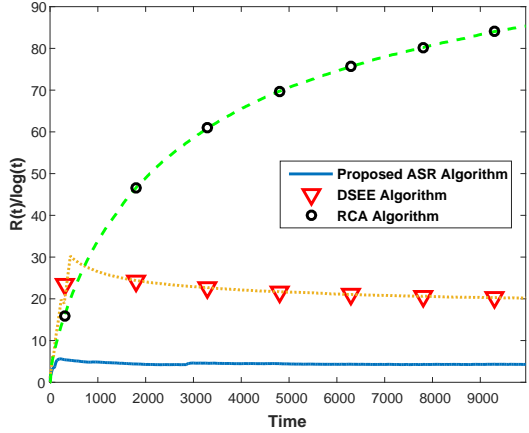,
width=0.45\textwidth}
\caption{The regret (normalized by $\log t$) under ASR, DSEE, and RCA as a function of time. Parameter setting (taken from \cite{Liu_2013_Learning}): 5 arms, each with two states: 0, 1. Transition probabilities for all arms to transit from 0 to 1 and from 1 to 0, respectively: $p_{01} = [0.1, 0.1, 0.5, 0.1, 0.1]$, $p_{10} = [0.2, 0.3, 0.1, 0.4, 0.5]$, reward for all arm at states 1, 0, respectively: $r_1 = [1, 1, 1, 1, 1]$, $r_0 =[0.1, 0.1, 0.1, 0.1, 0.1]$.
}
\label{fig:fig1}
\end{figure}

We first simulated the same scenario as in \cite[Figure 4]{Liu_2013_Learning}. The regret under the three algorithms is presented in Fig. \ref{fig:fig1}. It can be seen that ASR significantly outperforms both DSEE and RCA in this setting.\vspace{0.2cm}

\subsubsection{Increasing the system size}

\begin{figure}[htbp]
\centering \epsfig{file=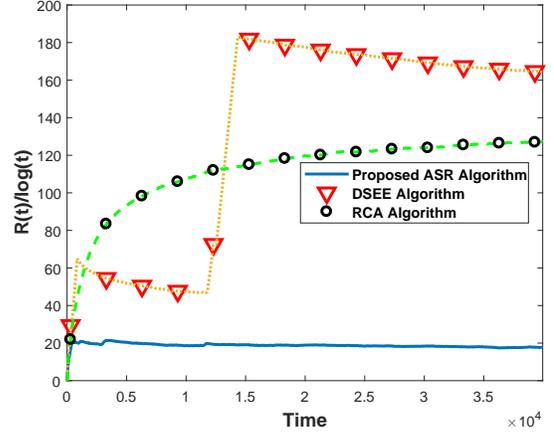,
width=0.45\textwidth}
\caption{The regret (normalized by $\log t$) under ASR, DSEE, and RCA as a function of time. Parameter setting: 10 arms, each with two states: 0, 1. Transition probabilities for all arms to transit from 0 to 1 and from 1 to 0, respectively: $p_{01} = [0.1, 0.1, 0.5, 0.1, 0.1, 0.2, 0.1, 0.2, 0.15, 0.25]$, $p_{10} = [0.2, 0.3, 0.1, 0.4, 0.5, 0.45, 0.35, 0.3, 0.5, 0.4]$, reward for all arm at states 1, 0, respectively: $r_1 = [1, 1, 1, 1, 1, 1, 1, 1, 1, 1]$, $r_0 =[0.1, 0.1, 0.1, 0.1, 0.1, 0.1, 0.1, 0.1, 0.1, 0.1]$.
}
\label{fig:fig2}
\end{figure}

Next, we are interested to examine the regret in a larger system. Therefore, in this scenario we increased the number of arms to $10$. As discussed in Section \ref{ssec:theoretical}, increasing the number of arms is expected to decrease the performance under DSEE since more arms are sampled by the worst exploration rate. Indeed, it can be seen in Fig. \ref{fig:fig2}, that RCA outperforms DSEE for the tested time horizon. It can be seen that ASR significantly outperforms both DSEE and RCA, due to the fact that each arm is played according to its unique exploration rate. \vspace{0.2cm}

\subsubsection{Decreasing the difference between the highest and the second highest rewards}

\begin{figure}[htbp]
\centering \epsfig{file=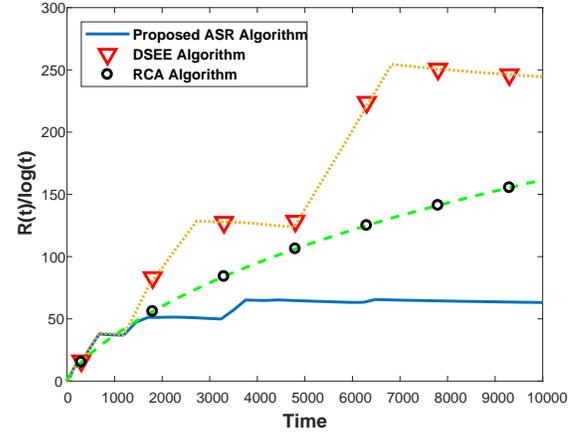,
width=0.45\textwidth}
\caption{The regret (normalized by $\log t$) under ASR, DSEE, and RCA as a function of time. Parameter setting: 5 arms, each with two states: 0, 1. Transition probabilities for all arms to transit from 0 to 1 and from 1 to 0, respectively: $p_{01} = [0.1, 0.8, 0.5, 0.1, 0.1]$, $p_{10} = [0.2, 0.2, 0.1, 0.4, 0.5]$, reward for all arm at states 1, 0, respectively: $r_1 = [1, 1, 1, 1, 1]$, $r_0 =[0.1, 0.1, 0.1, 0.1, 0.1]$.
}
\label{fig:fig3}
\end{figure}

In this section we are interested to examine the regret when the difference between the highest and the second highest reward means is relatively small. We simulated $5$ arms, and set the difference between the highest and the second highest reward means to $0.03$, which results in a high exploration rate used to distinguish between these two arms. As discussed in Section \ref{ssec:theoretical}, the DSEE algorithm explores all the arms with the high exploration rate, while RCA and ASR explore only these two arms using the high exploration rate. Indeed, as can be seen in Fig. \ref{fig:fig3}, this effect results in a high regret under DSEE as compared to RCA and ASR. The ASR algorithm significantly outperforms RCA and DSEE in this scenario again. \vspace{0.2cm}

\subsubsection{Increasing the state space}
\label{ssec:state_space}

\begin{figure}[htbp]
\centering \epsfig{file=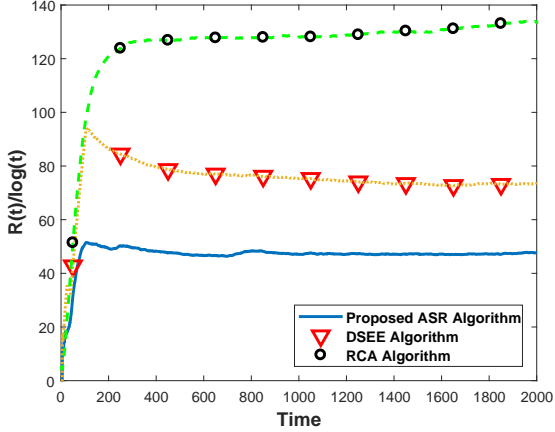,
width=0.45\textwidth}
\caption{The regret (normalized by $\log t$) under ASR, DSEE, and RCA as a function of time. Parameter setting: 5 arms, 20 states.
}
\label{fig:fig5}
\end{figure}

In this section we are interested in investigating the performance when the state space is relatively large. We simulated the same scenario as in \cite[Figure 7]{Liu_2013_Learning}. Specifically, we simulated a system with $5$ arms, and the number of states of each arm was set to $20$.
As shown in Fig. \ref{fig:fig5}, RCA performs poorly due to the long delay caused by the regenerative cycles. ASR outperforms both DSEE and RCA in this setting as well. \vspace{0.2cm}

\subsubsection{A case of bursty arms}

\begin{figure}[htbp]
\centering \epsfig{file=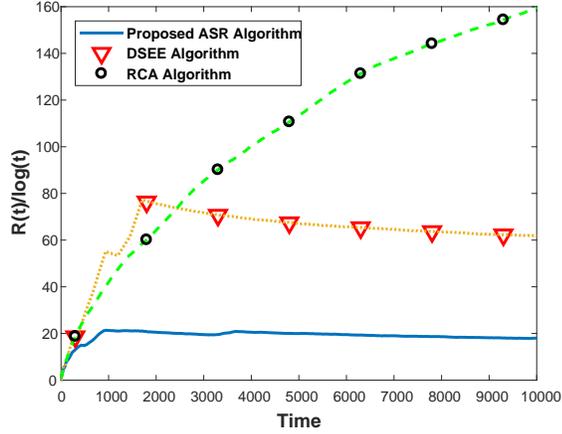,
width=0.45\textwidth}
\caption{The regret (normalized by $\log t$) under ASR, DSEE, and RCA as a function of time. Parameter setting: 5 arms, each with two states: 0, 1. Transition probabilities for all arms to transit from 0 to 1 and from 1 to 0, respectively: $p_{01} = [0.04, 0.05, 0.36, 0.05, 0.06]$, $p_{10} = [0.08, 0.15, 0.09, 0.05, 0.18]$, reward for all arm at states 1, 0, respectively: $r_1 = [1, 1, 1, 1, 1]$, $r_0 =[0.1, 0.1, 0.1, 0.1, 0.1]$.
}
\label{fig:fig4}
\end{figure}

Finally, we are interested to examine the performance in the case of bursty arms. Specifically, we set the probabilities of switching between states to be small. This setting is expected to significantly deteriorate performance under RCA due to the long delay caused by the regenerative cycles. As shown in Fig. \ref{fig:fig4}, the DSEE indeed outperforms RCA in this case. It can also be seen that ASR performs the best again.

\section{Conclusion}\label{sec:conclusion}

Inspired by recent developments of sequencing methods of exploration and exploitation epochs, we develop a novel algorithm that introduces the concept of adaptive sequencing rules for arm selection in RMAB problems, named Adaptive Sequencing Rules (ASR) algorithm. The arm sequencing rules are adaptive in the sense that they estimate the required exploration rate of each arm, and are updated dynamically with time, controlled by the random sample means in a closed loop manner. Significant performance gain of ASR algorithm over existing RCA and DSEE algorithms has been analyzed theoretically and numerically.

\section{Appendix}\label{sec:appendix}

In this Appendix, we prove the regret bound in (\ref{eq:regret}) shown in Theorem \ref{th:regret}.
The structure of the proof is as follows. Below, we define $T_1$, which is roughly speaking a random time by which the exploration rates for all arms are sufficiently close to the desired exploration rates needed for achieving the desired logarithmic regret bound (as shown later). We first show that the expectation of $T_1$ is bounded independent of $t$. Then, we will show that a logarithmic regret is obtained for all $t>T_1$, which yields the desired regret. \vspace{0.2cm}

\begin{definition}
Let $T_1$ be the smallest integer, such that for all $t \geq T_1$ the following holds: $\overline{D}_i \leq \widehat{D}_i(t)$ for all $i=1, ..., N$, and also $\widehat{D}_i(t) \leq \overline{D}_{i,max}$ for all $i\in\mathcal{K}$,
where \vspace{0.2cm}
\begin{center}
$\displaystyle\overline{D}_{i,max}\triangleq{\frac{4L}{(\mu_{\sigma(1)}-\mu_i)^2-2\epsilon}}.$ \vspace{0.2cm}
\end{center}
\end{definition}

In the next Lemma we show that the expected value of $T_1$ is bounded under the ASR algorithm. \vspace{0.2cm}

\textbf{Step 1: Bounding $E(T_1)$:}\vspace{0.2cm}\\
\begin{lemma}
\label{lemma:T1}
Assume that the ASR algorithm is implemented as described in Section \ref{sec:ASR}. Then, $E(T_1)<\infty$ is bounded independent of $t$. \vspace{0.1cm}
\end{lemma}

\textit{Proof}: $E(T_1)$ can be written as follows: \\
$ E[T_1]=\sum\limits_{n=1}^{\infty} n \cdot Pr\left(T_1= n \right)=
\sum\limits_{n=1}^{\infty}Pr\left(T_1\geq n \right)\\=
 \vspace{0.1cm} \hspace{0.3cm} \sum\limits_{n=1}^{\infty} Pr\{\bigcup\limits_{i\in\mathcal{K}}\bigcup\limits_{j=n}^{\infty}(\widehat{D}_i(j)<\overline{D}_i \mbox{\;or\;}
\widehat{D}_i(j)> \overline{D}_{i,max}) \mbox{\;or\;}\\
 \vspace{0.2cm} \hspace{3cm} \bigcup\limits_{i\nin\mathcal{K}}\bigcup\limits_{j=n}^{\infty}(\widehat{D}_i(j)<\overline{D}_i)
 \}
\\\leq
\vspace{0.2cm} \hspace{0.3cm} \sum\limits_{i\in\mathcal{K}}\sum\limits_{n=1}^{\infty} \sum\limits_{j=n}^{\infty}
Pr\{\widehat{D}_i(j)<\overline{D}_i \mbox{\;or\;}
\widehat{D}_i(j)> \overline{D}_{i,max}\} \\
\vspace{0.2cm} \hspace{0.3cm} +\sum\limits_{i\nin\mathcal{K}}\sum\limits_{n=1}^{\infty} \sum\limits_{j=n}^{\infty}
Pr\{\widehat{D}_i(j)<\overline{D}_i\} $\\
Note that if we show that
 \begin{align}
 Pr\{\widehat{D}_i(j)<\overline{D}_i \mbox{\;or\;}\widehat{D}_i(j)> \overline{D}_{i,max}\}\leq C\cdot j^{-(2+\delta)}
\end{align}
\vspace{0.1cm} for some constants $C>0, \delta > 0$ for all $i\in\mathcal{K}$ for all $j\geq n$, then we get: \vspace{0.2cm}\\
$ \displaystyle\sum\limits_{i\in\mathcal{K}}\sum\limits_{n=1}^{\infty} \sum\limits_{j=n}^{\infty}
Pr\{\widehat{D}_i(j)<\overline{D}_i \mbox{\;or\;}\widehat{D}_i(j)> \overline{D}_{i,max}\}
\\\leq
\vspace{0.1cm} \hspace{0.3cm} NC\left[\sum\limits_{j=1}^{\infty} j^{-(2+ \delta)}+\sum\limits_{n=2}^{\infty}\sum\limits_{j=n}^{\infty} j^{-(2+ \delta)}\right]
\\\leq
\vspace{0.1cm} \hspace{0.3cm} NC\left[\sum\limits_{j=1}^{\infty} j^{-(2+ \delta)}+\sum\limits_{n=2}^{\infty}\int\limits_{n-1}^{\infty} j^{-(2+ \delta)}dj\right]
\\=
\vspace{0.1cm} \hspace{0.3cm} NC\left[\sum\limits_{j=1}^{\infty} j^{-(2+ \delta)}+\frac{1}{1+\delta}\sum\limits_{n=2}^{\infty}(n-1)^{-(1+\delta)}\right]<\infty$,\vspace{0.2cm}\\
which is bounded independent of $t$. Similarly, showing that $Pr\{\widehat{D}_i(j)<\overline{D}_i\}\leq C\cdot j^{-(2+\delta)}$ for some constants $C, \delta > 0$ for all $i\nin\mathcal{K}$ for all $j\geq n$ completes the statement.
\vspace{0.2cm}

\noindent\textbf{Step 1.1: Proving the bound in (10):}\vspace{0.1cm}\\
\vspace{0.1cm} We denote $\tilde{s}_{\sigma(1)} \triangleq \max_{j} \{\tilde{s}_j(t)\}$. Then, we have, \\
$\vspace{0.2cm} Pr \{\widehat{D}_i(t)<\overline{D}_i \quad or
\quad \widehat{D}_i(t)> \overline{D}_{i,max}\}=\\
\vspace{0.3cm} \hspace{0.3cm} Pr \bigg\{\frac{4L}{\max\{\Delta,(\tilde{s}_{\sigma(1)}(t)-
\tilde{s}_i(t))^2-\epsilon\}} <
\frac{4L}{(\mu_{\sigma(1)}- \mu_i)^2} \\
\vspace{0.2cm} \hspace{0.7cm} \bigcup \frac{4L}{\max\{\Delta,(\tilde{s}_{\sigma(1)}(t)-
\tilde{s}_i(t))^2-\epsilon\}} >
\frac{4L}{(\mu_{\sigma(1)}- \mu_i)^2-2\epsilon}\bigg \} \\
\vspace{0.2cm} \hspace{0.1cm}=
Pr \bigg\{ \bigg[ \bigg((\tilde{s}_{\sigma(1)}(t)- \tilde{s}_i(t))^2- \epsilon> (\mu_{\sigma(1)}- \mu_i)^2  \\
\vspace{0.2cm} \hspace{1.7cm} \cap (\tilde{s}_{\sigma(1)}(t)- \tilde{s}_i(t))^2-\epsilon \geq \Delta \bigg) \\
 \vspace{0.2cm} \hspace{1.2cm} \bigcup \bigg( \Delta>(\mu_{\sigma(1)}- \mu_i)^2 \\
\vspace{0.2cm} \hspace{1.7cm} \cap (\tilde{s}_{\sigma(1)}(t)- \tilde{s}_i(t))^2-\epsilon<\Delta
 \bigg) \bigg] \\
\vspace{0.2cm} \hspace{0.8cm} \displaystyle \bigcup \bigg[ \bigg( (\tilde{s}_{\sigma(1)}(t)- \tilde{s}_i(t))^2- \epsilon <
(\mu_{\sigma(1)}- \mu_i)^2 -2\epsilon \\
\vspace{0.2cm} \hspace{1.7cm} \cap (\tilde{s}_{\sigma(1)}(t)- \tilde{s}_i(t))^2-\epsilon \geq \Delta \bigg) \\
 \vspace{0.2cm} \hspace{1.2cm} \textstyle \bigcup \bigg( \Delta<(\mu_{\sigma(1)}- \mu_i)^2-2\epsilon \\
 \vspace{0.2cm} \hspace{1.7cm} \cap (\tilde{s}_{\sigma(1)}(t)- \tilde{s}_i(t))^2-\epsilon<\Delta \bigg) \bigg] \bigg\} \\
    \vspace{0.2cm} \hspace{0.3cm} \leq Pr \bigg \{ \bigg [
(\tilde{s}_{\sigma(1)}(t)- \tilde{s}_i(t))^2- \epsilon> (\mu_{\sigma(1)}- \mu_i)^2  \\
 \vspace{0.2cm} \hspace{1.7cm} \bigcup \Delta>(\mu_{\sigma(1)}- \mu_i)^2  \bigg ]\\
  \vspace{0.2cm} \hspace{1.2cm} \displaystyle \bigcup \bigg[
 (\tilde{s}_{\sigma(1)}(t)- \tilde{s}_i(t))^2- \epsilon <
(\mu_{\sigma(1)}- \mu_i)^2 -2\epsilon \\
  \vspace{0.2cm} \hspace{1.7cm} \textstyle \bigcup (\tilde{s}_{\sigma(1)}(t)- \tilde{s}_i(t))^2-\epsilon<\Delta \bigg] \bigg\}.
$ \\
The probability for the second event on the RHS is zero, and the forth event lies inside the measure of the third event due to the fact that $i \in \mathcal{K}$.
Hence, \vspace{0.2cm}  \\
$\vspace{0.2cm} \hspace{0.3cm} Pr \{\widehat{D}_i(t)<\overline{D}_i \quad or \quad \widehat{D}_i(t)> \overline{D}_{i,max}\} \\
\vspace{0.2cm} \hspace{-0.2cm} \leq Pr\{
(\tilde{s}_{\sigma(1)}(t)- \tilde{s}_i(t))^2-(\mu_{\sigma(1)}- \mu_i)^2 > \epsilon \\
\vspace{0.4cm} \hspace{0.7cm} \bigcup
(\tilde{s}_{\sigma(1)}(t)- \tilde{s}_i(t))^2- (\mu_{\sigma(1)}- \mu_i)^2 < -\epsilon \} \\
\vspace{0.4cm} \hspace{-0.2cm} = Pr \{|(\tilde{s}_{\sigma(1)}(t)- \tilde{s}_i(t))^2-
(\mu_{\sigma(1)}- \mu_i)^2| > \epsilon \} \\
\vspace{0.2cm} \hspace{-0.2cm} = Pr \{|(\tilde{s}_{\sigma(1)}(t)- \tilde{s}_i(t))^2 - (\tilde{s}_{\sigma(1)}(t)- \tilde{s}_i(t))(\mu_{\sigma(1)}- \mu_i) \\
\vspace{0.4cm} \hspace{0.3cm} +(\tilde{s}_{\sigma(1)}(t)- \tilde{s}_i(t))(\mu_{\sigma(1)}- \mu_i)- (\mu_{\sigma(1)}- \mu_i)^2| > \epsilon \} \\
\vspace{0.2cm} \hspace{-0.2cm} = Pr \{|(\tilde{s}_{\sigma(1)}(t)- \tilde{s}_i(t)) [(\tilde{s}_{\sigma(1)}(t)- \tilde{s}_i(t))-(\mu_{\sigma(1)}- \mu_i)] \\
\vspace{0.4cm} \hspace{0.3cm} + (\mu_{\sigma(1)}- \mu_i)[(\tilde{s}_{\sigma(1)}(t)- \tilde{s}_i(t))-(\mu_{\sigma(1)}- \mu_i)]|> \epsilon \} \\
\vspace{0.2cm} \hspace{-0.2cm} \leq Pr\{|(\tilde{s}_{\sigma(1)}(t)- \tilde{s}_i(t)) [(\tilde{s}_{\sigma(1)}(t)- \tilde{s}_i(t))-(\mu_{\sigma(1)}- \mu_i)]| \\
 \vspace{0.0cm} \hspace{0.3cm} + |(\mu_{\sigma(1)}- \mu_i)[(\tilde{s}_{\sigma(1)}(t)- \tilde{s}_i(t))-(\mu_{\sigma(1)}- \mu_i)]|> \epsilon \} $

 \begin{align}
 \vspace{0.2cm} \hspace{-0.6cm} \leq & Pr\{|(\tilde{s}_{\sigma(1)}(t)- \tilde{s}_i(t)) [(\tilde{s}_{\sigma(1)}(t)- \tilde{s}_i(t))-(\mu_{\sigma(1)}- \mu_i)]|> \frac{\epsilon}{2} \}\nonumber \\
  \vspace{0.2cm} \hspace{-0.4cm} + &Pr\{|(\mu_{\sigma(1)}- \mu_i)[(\tilde{s}_{\sigma(1)}(t)- \tilde{s}_i(t))-(\mu_{\sigma(1)}- \mu_i)]|> \frac{\epsilon}{2} \}.
\end{align}
\vspace{0.2cm}

Now, we can observe that using concentration inequalities that bound the deviations of the sample mean estimates $\tilde{s}_{\sigma(1)}(t), \tilde{s}_i(t)$ from their true means $\mu_{\sigma(1)}, \mu_i$, respectively, will complete the statement. Hence, we next bound (11) in a tractable form so that we can use concentration inequalities by Lezaud's Lemma \cite{lezaud1998chernoff} for this.
We start by bounding the first term on the RHS of (11). For every $R>0$, we have: \vspace{0.2cm} \\
$
 \vspace{0.2cm} \hspace{-0.1cm} Pr\{|(\tilde{s}_{\sigma(1)}(t)- \tilde{s}_i(t)) [(\tilde{s}_{\sigma(1)}(t)- \tilde{s}_i(t))-(\mu_{\sigma(1)}- \mu_i)]|> \frac{\epsilon}{2} \} \\ \leq
 \vspace{0.2cm} Pr\{[|(\tilde{s}_{\sigma(1)}(t)- \tilde{s}_i(t))-(\mu_{\sigma(1)}- \mu_i)|>1] \bigcup \\
\vspace{0.2cm} \hspace{0.6cm} [|(\tilde{s}_{\sigma(1)}(t)- \tilde{s}_i(t))-(\mu_{\sigma(1)}- \mu_i)|> \frac{\epsilon}{2(R+1)}] \bigcup \\
\vspace{0.2cm} \hspace{0.6cm} [|(\mu_{\sigma(1)}- \mu_i)+1|>R] \}
 \\ \leq
\vspace{0.2cm} \hspace{0.0cm} Pr\{[|(\tilde{s}_{\sigma(1)}(t)- \tilde{s}_i(t))-(\mu_{\sigma(1)}- \mu_i)|>1]\\
\vspace{0.2cm} \hspace{0.6cm} +Pr\{|(\tilde{s}_{\sigma(1)}(t)- \tilde{s}_i(t))-(\mu_{\sigma(1)}- \mu_i)|> \frac{\epsilon}{2(R+1)}\} \\
\vspace{0.2cm} \hspace{0.6cm} +Pr\{|(\mu_{\sigma(1)}- \mu_i)+1|>R\}
 \\ \leq
\vspace{0.2cm} \hspace{0.1cm} 2 Pr \{ |(\tilde{s}_{\sigma(1)}(t)- \tilde{s}_i(t))-(\mu_{\sigma(1)}- \mu_i)|> \frac{\epsilon}{2(R+1)} \} \\
\vspace{0.2cm} \hspace{0.1cm} +Pr\{\mu_{\sigma(1)}+1>R \}. \\
$
\vspace{0.2cm} We choose $R=\mu_{\sigma(1)}+1$. Then, the second term is equal to 0. \vspace{0.2cm} We proceed with the first term:\\
$
\vspace{0.2cm} \hspace{0.1cm} 2\cdot Pr \{ |(\tilde{s}_{\sigma(1)}(t)- \tilde{s}_i(t))-(\mu_{\sigma(1)}- \mu_i)|> \frac{\epsilon}{2(R+1)} \} \\
\vspace{-0.1cm} \hspace{0.1cm} =2\cdot
Pr \{|(\tilde{s}_{\sigma(1)}(t)-\mu_{\sigma(1)})-(\tilde{s}_i(t)-\mu_i)|>\frac{\epsilon}{2(R+1)} \} $
\begin{align}
\vspace{0.2cm} \hspace{-2.7cm} \leq  2\cdot (Pr \{|\tilde{s}_{\sigma(1)}(t)-\mu_{\sigma(1)}|
>\frac{\epsilon}{4(R+1)} \}\nonumber \\
\vspace{-0.4cm} \hspace{-1.2cm} +Pr \{ |\tilde{s}_i(t)-\mu_i)|>\frac{\epsilon}{4(R+1)} \}).
\end{align}

\vspace{0.0cm} \hspace{-0.4cm} We next bound the second term on the RHS of (11). For every $R'>0$, we have: \vspace{0.2cm}\\
$\vspace{0.3cm} Pr\{|(\mu_{\sigma(1)}- \mu_i)[(\tilde{s}_{\sigma(1)}(t)- \tilde{s}_i(t))-(\mu_{\sigma(1)}- \mu_i)]|> \frac{\epsilon}{2} \} \\ \leq
\vspace{0.3cm} Pr \{\mu_{\sigma(1)}>R' \} \\
\vspace{0.3cm} +Pr\{|(\tilde{s}_{\sigma(1)}(t)- \tilde{s}_i(t))-(\mu_{\sigma(1)}- \mu_i)|>\frac{\epsilon}{2(R'+1)} \}.\\ $
\vspace{0.3cm} We now choose $R'=R=\mu_{\sigma(1)}+1$, so the first term is equal to 0. \vspace{0.3cm} We continue with the second term:\\
$\vspace{-0.1cm} Pr\{|(\tilde{s}_{\sigma(1)}(t)- \tilde{s}_i(t))-(\mu_{\sigma(1)}- \mu_i)|>\frac{\epsilon}{2(R'+1)} \} \leq$
\begin{align}
\vspace{0.3cm} \hspace{-3.4cm} Pr \{|\tilde{s}_{\sigma(1)}(t)-\mu_{\sigma(1)}|
>\frac{\epsilon}{4(R+1)} \} \\
\vspace{0.0cm} \hspace{-2.4cm}+ \nonumber Pr \{ |\tilde{s}_i(t)-\mu_i)|>\frac{\epsilon}{4(R+1)} \}.
\end{align}

\vspace{0.3cm}  \hspace{-0.4cm} By combining (12) and (13) we get:\\ \\
$\vspace{0.2cm} Pr \{\overline{D}_i(n)<\widehat{D}_i \quad or
\quad \overline{D}_i(n)> \overline{D}_{i,max}\} \leq \\
\vspace{0.2cm} 3\cdot (Pr \{|\tilde{s}_{\sigma(1)}(t)-\mu_{\sigma(1)}|
>\frac{\epsilon}{4(\mu_{\sigma(1)}+2)} \} + \\
\vspace{-0.1cm} \hspace{0.6cm} Pr \{ |\tilde{s}_i(t)-\mu_i)|>\frac{\epsilon}{4(\mu_{\sigma(1)}+2))} \}) $
\begin{align}
\vspace{0.2cm} \hspace{-0cm} \leq 6\cdot \max \left\{ Pr \left(|\tilde{s}_{\sigma(1)}(t)-\mu_{\sigma(1)}|
>\frac{\epsilon}{4(\mu_{\sigma(1)}+2)} \right) , \nonumber \right.\\\left.
\vspace{0.2cm} \hspace{-2cm} Pr \left( |\tilde{s}_i(t)-\mu_i)|>\frac{\epsilon}{4(\mu_{\sigma(1)}+2)} \right) \right\}.
\end{align}
\\
To complete the statement, we now ready to use Lezaud's results \cite{lezaud1998chernoff}, that bound the probability that a Markov chain will deviate from its stationary distribution:\\
\begin{lemma}[\cite{lezaud1998chernoff}] Consider a finite-state, irreducible Markov chain $\{X_t\}_{t \geq 1}$ with state space $S$, matrix of transition probabilities $P$, an initial distribution $q$,  and stationary distribution $\pi$. Let
$N_\textbf{q}=\left \| (\frac{q_x}{\pi_x}, x \in S) \right \|_2 $.
Let $\widehat{P}=P'P $ be the multiplicative symmetrization of $P$ where $P'$ is the adjoint of $P$ on $l_2(\pi)$. Let $\epsilon= 1-\lambda_2$, where $\lambda_2$ is the second largest eigenvalue of the matrix $P'$. $\epsilon$ will be referred to as the eigenvalue gap of $P'$. Let $f:S \rightarrow \mathcal{R}$ be such that $\sum\limits_{y \in S} \pi_yf(y)=0, \quad \|f\|_2 \leq 1$ and $0 \leq \|f\|_2^2 \leq 1$ if $P'$ is irreducible. Then, for any positive integer $n$ and all $0<\lambda\leq 1$, we have:\\
\begin{center}
$P \displaystyle \left(\frac{\sum\limits_{t=1} ^{n}f(X_t)}{n} \geq \lambda      \right) \leq N_\textbf{q}$ exp $[-\frac{n \lambda^2 \epsilon}{12}].$ \vspace{0.2cm}
\end{center}
\end{lemma}
Consider an initial distribution $\textbf{q}^i$ for the $i$th arm. We have:
\begin{center}
$ \displaystyle \left \| (\frac{q_i^s}{\pi_i^s}, s \in S^i) \right \|_2 \leq
  \sum\limits_{s \in S^i} \left \| \frac{q_i^s}{\pi_i^s} \right \|_2 \leq \frac{1}{\pi_{min}} $
\end{center}

Next, let $v_i(t)\triangleq|\mathcal{V}_i(t)|$ be the size of the set of all time indices during SB2 sub-blocks up to time $t$. Before applying Lezaud's bound, we pay attention for the following:
(i) The sample means $\tilde{s}_i(t)$ are calculated only from measurements in the set $\mathcal{V}_i$. As described in Section \ref{sec:ASR}, each interval in $\mathcal{V}_i$ starts from the last state that was observed in the previous interval. Therefore, cascading these intervals forms a sample path which is equivalent to a sample path generated by continuously sampling the Markov chain. Hence, we can apply Lezaud's bound to upper bound (14).
(ii) By the construction of the algorithm, (7) ensures that once exploitation epochs are executed (which are deterministic), the event $v_i(t)\geq\frac{(2+\delta)}{I}\log(t)$ for $\delta>0$ arbitrarily small surely occurs\footnote{We point out that a precise statement requires to set $(2+2\delta)$ in (7) and the statement holds for all $t>D$, where $D$ is a finite deterministic value. However, since $\delta>0$ is arbitrarily small and is not a design parameter, we do not present it explicitly when describing the algorithm to simplify the presentation.}. During exploration epochs, the randomness of SB1 (say for arm $r\neq i$) affects $v_i(t)$ since SB1 can be very long (with small probability) and then $v_i(t)\geq\frac{(2+\delta)}{I}\log(t)$ might not hold until the end of the epoch once the algorithm corrects the exploration gap by condition (7). Therefore, we define $E_i(t)$ as the event when all SB1 epochs that have been executed by time $t$ are smaller than $\delta\cdot t$. When event $E_i(t)$ occurs we have $v_i(t)\geq\frac{(2+\delta)}{I}\log(t)$ (for all $t>D$, for a sufficiently large finite deterministic value $D$). Then, for all $i$, we have: \vspace{0.2cm}\\
$\vspace{0.3cm}Pr \{ |\tilde{s}_i(t) -\mu_i| > \frac{\epsilon}{4(\mu_{\sigma(1)}+2)} \}\\=
\vspace{0.3cm} Pr \{ |\tilde{s}_i(t) -\mu_i| > \frac{\epsilon}{4(\mu_{\sigma(1)}+2)}, \mbox{\;$E_i(t)$ occurs}\}\\+
\vspace{0.3cm} Pr \{ |\tilde{s}_i(t) -\mu_i| > \frac{\epsilon}{4(\mu_{\sigma(1)}+2)}, \mbox{\;$E_i(t)$ does not occur}\}$
\begin{align}
 &\leq Pr \{ |\tilde{s}_i(t) -\mu_i| > \frac{\epsilon}{4(\mu_{\sigma(1)}+2)}, \mbox{\;$E_i(t)$ occurs}\}\\
 &+Pr\{\mbox{\;$E_i(t)$ does not occur}\}
\end{align}

Next, we upper bound both terms in (14) by bounding (15) and (16): \vspace{0.3cm}\\
$\vspace{0.3cm} Pr \{ |\tilde{s}_i(t) -\mu_i| > \frac{\epsilon}{4(\mu_{\sigma(1)}+2)}, \mbox{\;$E_i(t)$ occurs}\} \\ \leq
\vspace{0.3cm} Pr \{ |\tilde{s}_i(t) -\mu_i| > \frac{\epsilon}{4(\mu_{\sigma(1)}+2)}, \mbox{\;$E_i(t)$ occurs} \} \\
$
\vspace{0.3cm} We define $O_i^s(t)$ as the number of occurrences of state s on arm i up to time t, and we first look at: \vspace{0.3cm}\\
$\vspace{0.3cm} Pr \{ \tilde{s}_i(t) -\mu_i > \frac{\epsilon}{4(\mu_{\sigma(1)}+2)}, E_i(t)\} \\ =
\vspace{0.3cm} Pr \{\sum\limits_{s \in S^i} s \cdot O_i^s(t)-v^i(t) \sum\limits_{s \in S^i} s \cdot \pi_i^s > \frac{v_i(t)\cdot \epsilon}{4(\mu_{\sigma(1)}+2)}, E_i(t) \} \\ =
\vspace{0.3cm} \hspace{0.0cm} Pr \{\sum\limits_{s \in S^i} (s \cdot O_i^s(t)-v^i(t)s \cdot \pi_i^s) > \frac{v_i(t)\cdot \epsilon}{4(\mu_{\sigma(1)}+2)}, E_i(t) \} \\ \leq
\vspace{0.3cm} \sum\limits_{s \in S^i} Pr \{s \cdot O_i^s(t)-v^i(t)s \cdot \pi_i^s) > \frac{v_i(t)\cdot \epsilon}{4(\mu_{\sigma(1)}+2) |S^i|}, E_i(t) \} \\ =
\vspace{0.3cm} \sum\limits_{s \in S^i} Pr \{ O_i^s(t)-v^i(t) \cdot \pi_i^s) > \frac{v_i(t)\cdot \epsilon}{4(\mu_{\sigma(1)}+2) |S^i| \cdot s}, E_i(t)\} \\ =
\vspace{0.3cm} \sum\limits_{s \in S^i} Pr \{ \frac{\sum \limits_{n=1}^t \textbf{1} (s_i(n)=s)-v_i(t) \pi_i^s } {\hat{\pi}_i^s \cdot v_i(t)} > \frac{v_i(t)\cdot \epsilon}{4(\mu_{\sigma(1)}+2) |S^i| \cdot s \hat{\pi}_i^s }, E_i(t)\} \\ \leq
\vspace{0.3cm} \hspace{0.3cm} |S^i| \cdot N_\textbf{q}^{(i)}$ exp $(-v_i(t) \cdot \frac{\epsilon^2}{16(\mu_{\sigma(1)}+2)^2 \cdot s^2 \cdot |S^i|^2 \cdot \hat{\pi}_i^2} \cdot \frac{(1-\lambda_i)}{12}) \\ $
and due to $E_i(t)$: $v_i(t)> \frac{2+\delta}{I} \cdot log(t)$ so we have: \vspace{0.3cm} \\
$\vspace{0.3cm} Pr \{ \tilde{s}_i(t) -\mu_i > \frac{\epsilon}{4(\mu_{\sigma(1)}+2)}, E_i(t)\} \\ \leq
\vspace{0.3cm} \hspace{0.3cm}\frac{|S_{max}|}{\pi_{min}}$ exp $(- \frac{(2+ \delta)}{I} \cdot \frac{\epsilon^2 \cdot (1-\lambda_i)}{12 \cdot 16(\mu_{\sigma(1)}+2)^2\cdot s^2 \cdot |S^i|^2 \cdot \hat{\pi}_i^2} \cdot \log(t))\\=
\vspace{0.3cm} \hspace{0.3cm} \frac{|S_{max}|}{\pi_{min}}$exp$(- \frac{(2+ \delta) 192 (r_{max}+2)^2 \cdot S_{max}^2 r_{max}^2 \hat{\pi}_{max}^2}{\epsilon^2(1-\lambda_{max})}   \cdot \\
\vspace{0.3cm} \hspace{2cm}   \frac{\epsilon^2 \cdot (1-\lambda_i)}{12 \cdot 16(\mu_{\sigma(1)}+2)^2\cdot s^2 \cdot |S^i|^2 \cdot \hat{\pi}_i^2} \cdot \log(t))\\ \leq
\vspace{0.3cm} \hspace{0.3cm} \frac{|S_{max}|}{\pi_{min}} \cdot e^{-(2+\delta)\cdot \log(t)}=\frac{|S_{max}|}{\pi_{min}} \cdot t^{-(2+\delta)} \leq \frac{|S_{max}|}{\pi_{min}} \cdot t^{-(2+\delta)}$.\vspace{0.2cm}\\
for some $\delta>0$ arbitrarily small. \\
By applying Lemma 3 to $-f$ we get the same bound on
$P  \left(\frac{\sum\limits_{t=1} ^{n}f(X_t)}{n} \leq -\lambda \right) $
, and thus we get the bound for (15).
Next, we upper bound (16). When event $E_i(t)$ does not occur, there exists an SB1 epoch (i.e., hitting time) which is greater than $\delta\cdot t$. Therefore, there exist $C, \gamma, C_1>0$, such that $Pr\{\mbox{\;$E_i(t)$ does not occur}\}\leq C_1 t\cdot e^{-\gamma t}\leq Ct^{-(2+\delta)}$, which completes (10). Showing that $Pr\{\widehat{D}_i(j)<\overline{D}_i\}\leq j^{-(2+\delta)}$ for some $\delta > 0$ for all $i\nin\mathcal{K}$ for all $j\geq n$ follows similar steps as we showed by handling $\widehat{D}_i(j)<\overline{D}_i$ when proving (10). Thus, Lemma \ref{lemma:T1} follows.
\hfill $\square$ \vspace{0.2cm}

\noindent\textbf{Step 2: Continuing proving the Theorem using Lemma \ref{lemma:T1}:}\vspace{0.2cm}

The regret can be written as follows:\\
$\vspace{0.3cm} r_\Phi(t)= E[\sum\limits_{\tau=1}^{t}\mu_{\sigma(1)} -\sum\limits_{i=1}^{N}\sum\limits_{n=1}^{T_i(t)}s_i(t_i(n))] \\=
\vspace{0.3cm} \hspace{0.3cm} E[(\sum\limits_{\tau=1}^{T_1}\mu_{\sigma(1)}+ \sum\limits_{\tau=T_1+1}^{t}\mu_{\sigma(1)})-\\
\vspace{0.0cm} \hspace{0.6cm}
(\sum\limits_{i=1}^{N}\sum\limits_{n=1}^{T_i(T_1)}s_i(t_i(n))+\sum\limits_{i=1}^{N}\sum\limits_{n=T_i(T_1+1)}^{T_i(t)}s_i(t_i(n)))]=$\\
\begin{align}
&\hspace{-0.6cm}\mu_{\sigma(1)}E[T_1]-E[\sum\limits_{i=1}^{N}\sum\limits_{n=1}^{T_i(T_1)}s_i(t_i(n))] \\+
&E[\sum\limits_{\tau=T_1+1}^{t}\mu_{\sigma(1)})-\sum\limits_{i=1}^{N}\sum\limits_{n=T_i(T_1+1)}^{T_i(t)}s_i(t_i(n)))].
\end{align}
By applying Lemma \ref{lemma:T1}, we obtain that (17) is bounded independent of $t$:
\beq
\mu_{\sigma(1)}E[T_1]-E[\sum\limits_{i=1}^{N}\sum\limits_{n=1}^{T_i(T_1)}s_i(t_i(n))]\leq\mu_{\sigma(1)}E[T_1]=O(1),
\eeq
which results in the additional constant term $O(1)$ in the regret bound in (8) which is independent of $t$.

Next, we upper bound (18). Note that for all $t>T_1$, we have: \\
\begin{align}
 \overline{D}_i \leq \widehat{D}_i(t) \leq \overline{D}_{i,max},
\end{align} \\
for all $i\in\mathcal{K}$, and we have the LHS of the inequality for $i\nin\mathcal{K}$.
For convenience, we will develop (18) between $\tau=1$ and $t$ with (20) (and the LHS for $i\nin\mathcal{K}$) holds for all $1\leq\tau\leq t$, which upper bounds (18) between $\tau=T_1$ and $t$:\\
$\vspace{0.3cm} E[\sum\limits_{\tau=T_1+1}^{t}\mu_{\sigma(1)})-\sum\limits_{i=1}^{N}\sum\limits_{n=T_i(T_1+1)}^{T_i(t)}s_i(t_i(n)))] \leq \\
 \hspace{0.0cm} E[\sum\limits_{\tau=1}^{t}\mu_{\sigma(1)})-\sum\limits_{i=1}^{N}\sum\limits_{n=1}^{T_i(t)}s_i(t_i(n)))]=$ \\
\begin{align}
 t \cdot \mu_{\sigma(1)}-E[\sum\limits_{i=1}^{N}\sum\limits_{n=1}^{T_i(t)}s_i(t_i(n))].
\end{align} \vspace{0.2cm}

\noindent\textbf{Step 2.1: Showing a logarithmic order of (21):}\vspace{0.2cm}

\vspace{0.3cm} We next show that (21) has a logarithmic order with $t$:\\
\begin{center}
$\vspace{0.1cm} t \cdot \mu_{\sigma(1)}-E[\sum\limits_{i=1}^{N}\sum\limits_{n=1}^{T_i(t)}s_i(t_i(n))]$
\end{center}
\begin{align}
\vspace{0.3cm}  =\sum\limits_{i=1}^{N} &  \bigg[\mu_i E[T_i(t)]-E
 [\sum \limits_{n=1}^{T_i(t)}s_i(t_i(n))]\bigg ]\\+ \nonumber
 & \bigg[ t\cdot \mu_{\sigma(1)}-
\sum \limits_{i=1}^{N}\mu_i E[T_i(t)] \bigg].
\end{align} \\
We next show that both terms in (22) have a logarithmic order with $t$, which results in a logarithmic order with $t$ for the regret. We will divide the regret of each term for the exploration and exploitation epochs.
The first term of (22) can be viewed as the regret caused by arm switchings, and the second term can be viewed as the regret caused by choosing a sub-optimal arm.
We first bound the first term using the following lemma:\\
\begin{lemma}[\cite{Anantharam_1987_Asymptotically}]
\label{lemma:ananthram}
Consider an irreducible, aperiodic Markov chain with state space $S$, matrix of transition probabilities $P$, an initial distribution $\overrightarrow{q}$ which is positive in all states, and stationary distribution $\overrightarrow{\pi} (\pi_s$ is the stationary probability of state s). The state (reward) at time $t$ is denoted by $s(t)$. Let $\mu$ denote the mean reward. If we play the chain for an arbitrary time $T$, then there exists a value $A_p \leq (\min_{s \in S}\pi_s)^-1 \sum \limits_{s \in S} s$, such that: $E[\sum\limits_{t=1}^{T}s(t)-\mu T] \leq A_p$.
\end{lemma}

Lemma \ref{lemma:ananthram} addresses the difference between the expected reward obtained by playing an arm for time $T$, and $T\mu$. When applying the lemma to our case, it shows that the upper bound for the regret caused by each arm switching is a constant independent of the amount of time we played the arm in each epoch.
For the exploration epochs, we upper bound the number of exploration epochs $n_O^i$ for each arm (say $i$) by time $t$. If the player has started the $n^{th}$ exploration epoch, we have by (7) and the fact that $t\geq T_1$:
\begin{center}
$\vspace{0.3cm} \sum \limits_{n=1}^{n_O^i} 4^{n-1} =\frac{1}{3}(4^{n_O^i}-1) \leq A_i \cdot \log (t) $,
\end{center}
where
\begin{center}
$A_i\triangleq
\left\{ \begin{matrix}
\max\{2/I\;,\;\overline{D}_{i,max}\} \;, & \mbox{if $i\in\mathcal{K}$}    \vspace{0.2cm}\\
\max \{2/I\;,\;4L/\Delta\} \;, & \mbox{if $i\nin\mathcal{K}$}
\end{matrix} \right.  $\;.
\end{center}
Hence,
\beq
n_O^i(t) \leq \lfloor \log_4(3A_i\log(t)+1) \rfloor +1.
\eeq
 For the exploitation epochs, by time $t$, at most $(t-N)$ time slots have been spent on exploitation epochs (if we only performed a single exploration epoch for every arm with one play in the beginning of the algorithm). Thus, we have:
\begin{center}
$\vspace{0.3cm} \sum \limits_{n=1}^{n_I} 2\cdot 4^{n-1} \leq (t-N)$,
\end{center}
which implies
\begin{center}
$\frac{2}{3}(4^{n_I}-1) \leq (t-N)$.
\end{center}
Hence,
\beq
n_I \leq \lceil \log_4(\frac{3}{2}(t-N)+1) \rceil.
\eeq
As a result, there are a logarithmic number of exploitation epochs with time, each applies an arm switching. Therefore, the total regret of the first term in (22) is upper bounded by:
\beq
\bea{l}
\displaystyle\sum\limits_{i=1}^{N}\bigg[\mu_i E[T_i(t)]-E[\sum \limits_{n=1}^{T_i(t)}s_i(t_i(n))]\bigg ]\leq\vspace{0.2cm}\\ \hspace{2cm}
\displaystyle \hspace{-0.7cm} A_{max}\cdot \bigg( \sum\limits_{i=1}^{N} (\lfloor \log_4(3A_i\log(t)+1) \rfloor +1) \\ \hspace{2cm}
 \hspace{0.6cm} + \lceil \log_4(\frac{3}{2}(t-N)+1) \rceil \bigg),
\ena
\eeq
which coincides with the first and third term on the RHS of (8).
Next, we show that the second term in (22) has a logarithmic order with time. The approach here is to show that for every bad arm $i$, $E[T^i(t)]$ has a logarithmic order with time. Let $T^i_O(t)$, and $T^i_I(t)$, denote the time spent on arm $i$ in exploration and exploitation epochs, respectively, by time $t$. Thus,
\begin{center}
$T^i(t)=T^i_O(t)+T^i_I(t)$.
\end{center}
We will show that both $E[T^i_O(t)]$ and $E[T^i_I(t)]$ have a logarithmic order with time.

We start by handling the exploration epochs. Note that exploration epoch $n_O^i$ for arm $i$ consists of the time until the last state observed at the $(n_O^i-1)^{th}$ exploration epoch $\gamma^i(n_O^i-1)$ is observed again (i.e., SB1 sub-block), and another $4^{n_O^i}$ time slots.
The bound in (23) still holds,
thus, the time spent by time $t$ in exploration epochs for arm $i$ is bounded by: \vspace{0.3cm} \\
$\vspace{0.3cm}E[T_O^i(t)] \leq \sum \limits_{n=0}^{n_O^i-1}(4^n+M^i_{max})= \\
\vspace{0.3cm} \hspace{0.3cm}\frac{1}{3}(4^{n_O^i(t)}-1)+ M^i_{max} \cdot n_O^i(t) \leq \\
\vspace{0.3cm} \hspace{0.3cm} \frac{1}{3} [4(3A_i\cdot \log(t)+1)-1]+
M^i_{max} \cdot \log_4(3A_i\log(t)+1),\\
$
and the regret caused by playing bad arms in exploration epochs by time $t$ is bounded by: \vspace{0.3cm} \\
$\vspace{0.3cm} \sum \limits_{i=1}^N (\mu_{\sigma(1)}-\mu_i) \cdot
\bigg[\frac{1}{3} [4(3A_i\cdot \log(t)+1)-1]\\
\vspace{0.3cm} \hspace{3cm}  +M^i_{max} \cdot \log_4(3A_i\log(t)+1) \bigg] \\
\vspace{0.3cm} =\sum \limits_{i=1}^N (\mu_{\sigma(1)}-\mu_i) \cdot
4A_i\cdot \log(t)\\
\vspace{0.3cm} \hspace{0.6cm} +\sum \limits_{i=1}^N (\mu_{\sigma(1)}-\mu_i) \cdot [M^i_{max} \cdot \log_4(3A_i\log(t)+1)+1], \\
$
which coincides with the second and third terms on the RHS of (8) by simple algebraic manipulations.

Next, in order to bound the regret caused by playing bad arms in exploitation epochs, we define $Pr[i,n]$ as the probability that a sub-optimal arm $i$ is played in the $n^{th}$ exploitation epoch. From the upper bound on the number of the exploitation epochs given in (24), we thus have: \vspace{0.3cm} \\
$\vspace{0.3cm} E[T^i_I(t)]= \sum \limits_{n=1}^{n_I} 2\cdot 4^{n-1} \cdot Pr[i,n] \\
\vspace{0.3cm} \hspace{0.6cm} \leq \sum \limits_{n=1}^{\lceil \log_4(\frac{3}{2}(t-N)+1) \rceil} 2\cdot 4^{n-1} \cdot Pr[i,n]
$
\begin{align}
\leq\sum \limits_{n=1}^{\lceil \log_4(\frac{3}{2}(t-N)+1) \rceil} 3t_n \cdot Pr[i,n],
\end{align} \\
where $t_n$ denotes the starting time of the $n^{th}$ exploitation epoch and (26) follows from the fact that $t_n \geq \frac{2}{3} 4^{n-1} $.
Note that it suffices to show that $Pr[i,n]$ has an order of $t_n^{-1}$ so as to obtain a logarithmic order with time for the summation in (26).

We next bound $Pr[i,n]$. We define $C_{t,w}= \sqrt{L \log(t)/w }$, and let $w_i$ and $w_{\sigma(1)}$ denote, respectively, the number of plays on arm $i$ and the best arm by time $t_n$.
Recall that $\overline{s}_i$ denotes the sample mean of arm $i$ computed from samples from SB2 and exploitation epochs. Thus,
\beq
\bea{l}
Pr[i,n]= Pr \{\overline{s}_i(t_n) \geq \overline{s}_{\sigma(1)}(t_n) \}\vspace{0.2cm}\\ \hspace{0.5cm}
\leq Pr\{\overline{s}_{\sigma(1)}(t_n) \leq \mu_{\sigma(1)}- C_{t_n,w_{\sigma(1)}} \} \vspace{0.2cm}\\ \hspace{1cm}
+Pr\{\overline{s}_i(t_n) \geq \mu_i+ C_{t_n,w_i} \} \vspace{0.2cm}\\ \hspace{1cm}
+Pr\{ \mu_{\sigma(1)} <\mu_i+ C_{t_n,w_i}+ C_{t_n,w_{\sigma(1)}} \}.
\ena
\eeq

We first show that the third term in (27) is zero. Note that from (7) we have:
\begin{center}
$\vspace{0.3cm} w_i> \max \{\widehat{D}_i(t),\frac{2}{I} \} \cdot \log t_n$,
\end{center}
and from (20) and the fact that $\overline{D}_i \leq \overline{D}_{\sigma(1)} $, we have:
\begin{center}
$\min\left\{w_{\sigma(1)} , w_i\right\} \geq \overline{D}_i \cdot \log t_n$.\vspace{0.2cm}
\end{center}
As a result,
\begin{center}
$\bea{l}
Pr \{\mu_{\sigma(1)} <\mu_i+ C_{t_n,w_i}+ C_{t_n,w_{\sigma(1)}} \}\vspace{0.2cm}\\
=Pr\{ \mu_{\sigma(1)} - \mu_i < \sqrt{\frac{L \log t_n}{w_i}}+\sqrt{\frac{L \log t_n}{w_{\sigma(1)}}}\}\vspace{0.2cm}\\
\leq Pr\{ \mu_{\sigma(1)} - \mu_i < 2\sqrt{\frac{L \log t_n}{\min\left\{w_{\sigma(1)} , w_i\right\}}}\}\vspace{0.2cm}\\
=Pr\{ (\mu_{\sigma(1)} - \mu_i)^2< \frac{4L \log t_n}{\min\left\{w_{\sigma(1)} , w_i\right\}} \} \vspace{0.2cm}\\
= Pr \{\min\left\{w_{\sigma(1)} , w_i\right\} < \overline{D}_i \cdot \log t_n \}=0.
\ena$
\end{center}
Therefore, we can rewrite (27) as follows:
\beq
\bea{l}
Pr[i,n] \leq Pr\{\overline{s}_{\sigma(1)}(t_n) \leq \mu_{\sigma(1)}- C_{t_n,w_{\sigma(1)}} \} \vspace{0.2cm}\\ \hspace{2cm}
                + Pr\{\overline{s}_i(t_n) \geq \mu_i+ C_{t_n,w_i} \}.
\ena
\eeq

Next, we bound both terms on the RHS of (28). For the second term, the event: $\overline{s}_i(t_n) \geq \mu_i+ C_{t_n,w_i}$ is equivalent to:
\begin{center}
$\vspace{0.3cm} w_i\overline{s}_i(t_n) \geq w_i\mu_i+ \sqrt{L w_i \log t_n}$.
\end{center}
The inequality implies that the sample mean for arm $i$ largely deviates from its expected value. This event implies that the sample mean from at least one epoch largely deviates from the true mean. Note that the total plays on an arm consists of multiple contiguous segments of the Markov sample path, each in a different epoch. The possible values for the number of plays in the exploitation epochs are $2\times4^n$. The possible values of the numbers of plays on an arm in the exploration epochs at SB2 blocks are $4^n$. Consequently, we can write the time $w_i$ spent on arm $i$ as epochs with lengths $2^{n_1^i-1},2^{n_2^i-1}, \ldots , 2^{n_K^i-1}$, where $K$ is the number of epochs, and $n_j$ is distinct ($n_1<n_2< \ldots < n_K$). As a result, $w_i= \sum \limits_{j=1}^K 2^{n_j-1}$ and
$\sqrt{w_i} \geq \sum \limits_{j=1}^K (\sqrt{2}-1) \sqrt{2^{n_j-1}}$.
Let $R_i(2^{n_j-1})$ denote the total reward obtained during the $j^{th}$ segment. Using similar bounds as in \cite{Liu_2013_Learning}, we can show that the second term of (28) is bounded by: \vspace{0.3cm} \\
$\vspace{0.3cm} Pr\{ w_i\overline{s}_i(t_n) \geq w_i\mu_i+ \sqrt{L w_i \log t_n} \} \\
\vspace{0.3cm} \leq Pr \bigg \{\sum \limits_{j=1}^K R_i(2^{n_j-1})  \\
\vspace{0.3cm} \hspace{0.8cm} \geq\mu_i  \sum \limits_{j=1}^K 2^{n_j-1}+
\sqrt{L \log(t_n)} (\sqrt{2}-1 )\sum \limits_{j=1}^K \sqrt{2^{n_j-1}}
\bigg \} \\
\vspace{0.3cm} =Pr \bigg \{\sum \limits_{j=1}^K  \bigg (R_i(2^{n_j-1})
-\mu_i 2^{n_j-1}  \\
\vspace{0.3cm} \hspace{0.8cm} -\sqrt{L \log(t_n) }(\sqrt{2}-1 ) \sqrt{2^{n_j-1}} \bigg ) \geq 0 \bigg \} \\
\vspace{0.3cm} \displaystyle \leq\sum \limits_{j=1}^K Pr \bigg \{R_i(2^{n_j-1}) -\mu_i 2^{n_j-1}  \\
\vspace{0.3cm} \hspace{1cm} \geq(\sqrt{2}-1 ) \sqrt{L \log(t_n) 2^{n_j-1}}  \bigg \}. \\
$
The probability
\begin{center}
$Pr \bigg \{R_i(2^{n_j-1}) -\mu_i 2^{n_j-1} \geq (\sqrt{2}-1 ) \sqrt{L \log(t_n) 2^{n_j-1}}  \bigg \}$
\end{center}
is the probability for the event that the sum of rewards during a period of time with length $2^{n_j-1}$ for arm $i$ largely deviates from $\mu_i 2^{n_j-1}$. It can be written in terms of the numbers of occurrences of states. Let $O_i^s(j)$ denote the number of occurrences of state $s$ on arm $i$ in the $j^{th}$ segment.
Then, we have:
\begin{center}
$\bea{l}
Pr \bigg \{R_i(2^{n_j-1}) -\mu_i 2^{n_j-1} \geq (\sqrt{2}-1 ) \sqrt{L \log(t_n) 2^{n_j-1}}  \bigg \}\\
=Pr \bigg \{ \sum \limits_{s \in S}(sO_i^s(2^{n_j-1}) -s2^{n_j-1} \pi_s^i  \vspace{0.2cm}\\
\hspace{3cm} \geq(\sqrt{2}-1 ) \sqrt{L \log(t_n) 2^{n_j-1}} \bigg \},
\ena$
\end{center}
which after simple algebraic manipulations is upper bounded by: \vspace{0.3cm} \\
$\vspace{0.3cm} \displaystyle \sum \limits_{s \in S} Pr \bigg \{
O_i^s(2^{n_j-1})-2^{n_j-1} \pi_s^i  \\
\vspace{0.3cm}  \hspace{1cm} \geq(\sqrt{2}-1 ) \sqrt{L \log(t_n) 2^{n_j-1}} \left( \frac{1}{\sum \limits_{s\in S_i}s} \right) \bigg \}
$ \\

As a result, the event in which the sample mean largely deviates from the true mean implies that at least one state is visited much more often than predicted by its stationary probability. For bounding this probability, we use the bound by Gillman presented in the lemma below. \vspace{0.2cm}

\begin{lemma}[\cite{gillman1998chernoff}] Consider a finite state, irreducible, aperiodic, and reversible Markov chain with state space $S$, matrix of transition probabilities $P$, and an initial distribution $\textbf{q}$. Let $N_{\textbf{q}}=|(\frac{q_x}{\pi_x}),x \in S|_2$. Let $\epsilon$ be the eigenvalue gap given by $1-\lambda_2 $ where $\lambda_2$ is the second largest eigenvalue of the matrix $P$. Let $A \subseteq S$ and $T_A(t)$ be the number of times that states in $A$ are visited by time $t$. Then, for any $\gamma \geq 0$ we have:
\begin{align}
Pr\{T_A(t)-t\pi_A \geq \gamma \} \leq (1+\frac{\gamma\epsilon}{10t})N_{\textbf{q}}e^{-\gamma^2\epsilon/20t}.\vspace{0.3cm}
\end{align}\vspace{0.1cm}
\end{lemma}
Using Gillman's bound, we have:\hspace{-0.2cm} \vspace{0.3cm} \\
$\vspace{0.3cm}Pr\{\overline{s}_i(t_n) \geq \mu_i+ \sqrt{Lw_i\log(t_n} \} \\\leq
\vspace{0.3cm} \displaystyle \sum \limits _{j=1}^K \bigg ( \displaystyle \sum \limits_{s \in S} Pr \bigg \{
O_i^s(2^{n_j-1})-2^{n_j-1} \pi_s^i \geq \\
\vspace{0.3cm}  \hspace{1cm} (\sqrt{2}-1 ) \sqrt{L \log(t_n) 2^{n_j-1}} \left( \frac{1}{\sum \limits_{S_i}s} \right) \bigg \} \bigg) \\
\leq K|S_i|N_{\textbf{q}_i}t_n^{-((3-2\sqrt{2})L\overline{\lambda}_i)/20(\sum
\limits_{s \in S_i}s)^2))} \vspace{0.2cm}\\
\vspace{0.5cm}\hspace{0.5cm} + |S_i|
\frac{\sqrt{2}\overline{\lambda}_i\sqrt{L\log t_n}}{10 \sum \limits_{s \in S_i}s}N_{\textbf{q}_i}t_n^{-(3-2\sqrt{2})\frac{L\overline{\lambda}_i}{20(\sum \limits_{s \in S_i}s)^2)}} \\\leq
\vspace{0.5cm} \left(\frac{1}{\log2}+ \frac{\sqrt{2}\overline{\lambda}_i\sqrt{L}} {10 \sum \limits_{s \in S_i}s}\right) \times \\
\vspace{0.5cm}\hspace{2cm} |S_i|N_{\textbf{q}_i}t_n^{
1/2-(3-2\sqrt{2})(L\overline{\lambda}_i/(20(\sum \limits_{s \in S_i}s)^2))}
$,\vspace{0.3cm} \\
and by using (1) we get: \vspace{0.3cm} \\
$\vspace{0.3cm} Pr\{\overline{s}_i(t_n) \geq \mu_i+ C_{t_n,w_i} \}$
\begin{align}
 \leq\left(\frac{1}{\log2}+ \frac{\sqrt{2}\overline{\lambda}_i\sqrt{L}} {10 \sum \limits_{s \in S_i}s}\right) \cdot |S_i|N_{\textbf{q}_i}t_n^{-1}.
\end{align} \\
Similarly, we have: \vspace{0.3cm} \\
$\vspace{0.3cm} Pr\{\overline{s}_{\sigma(1)}(t_n) \leq \mu_{\sigma(1)}- C_{t_n,w_{\sigma(1)}} \} $
\begin{align}
\leq\left(\frac{1}{\log2}+ \frac{\sqrt{2}\overline{\lambda}_{\sigma(1)}\sqrt{L}} {10 \sum \limits_{s \in S_{\sigma(1)}}s}\right) \times |S_{\sigma(1)}|N_{\textbf{q}_i}t_n^{-1}.
\end{align} \\
Using (26) and the above, we can finally upper bound the regret caused by playing bad arms in exploitation epochs by:
\begin{align}
3\lceil \log_4(\frac{3}{2}(t-N)+1) \rceil \frac{1}{\pi_{min}}
\bigg[\sum \limits_{i\neq \sigma(1)}^N(\mu_{\sigma(1)}-\mu_j) \nonumber \\
\sum \limits_{k=1,i} \left( \frac{1}{\log2}+ \frac{\sqrt{2}\overline{\lambda}_k\sqrt{L}} {10 \sum \limits_{s \in S_k}s} \right) |S_k| \bigg],
\end{align} \\
which completes the first term in (8), and thus completes the proof.

\bibliographystyle{ieeetr}

\end{document}